\documentclass[sigconf]{acmart}  

\usepackage{graphicx}
\usepackage{multirow} 

\usepackage{booktabs}
\usepackage{longtable}
\usepackage{geometry}

\AtBeginDocument{%
  }

\setcopyright{none}
\acmConference{}{}{}
\acmBooktitle{}
\acmPrice{}
\acmDOI{}

\acmConference[]{ACM Transactions on Intelligent Systems and Technology}{March 2025}{USA}
\acmISBN{}

\begin{document}

\title{LLMs for Explainable AI: A Comprehensive Survey}

\author{Ahsan Bilal, David Ebert, Beiyu Lin\textsuperscript{*}}
\affiliation{%
  \institution{University of Oklahoma}
  \city{Norman}
  \state{OK}
  \country{USA}
}
\email{{ahsan.bilal-1, ebert, beiyu.lin}@ou.edu}

\begin{abstract}
Large Language Models (LLMs) offer a promising approach to enhancing Explainable AI (XAI) by transforming complex machine learning outputs into easy-to-understand narratives, making model predictions more accessible to users, and helping bridge the gap between sophisticated model behavior and human interpretability. AI models, such as state-of-the-art neural networks and deep learning models, are often seen as "black boxes" due to a lack of transparency. As users cannot fully understand how the models reach conclusions, users have difficulty trusting decisions from AI models, which leads to less effective decision-making processes, reduced accountabilities, and unclear potential biases. A challenge arises in developing explainable AI (XAI) models to gain users' trust and provide insights into how models generate their outputs. With the development of Large Language Models, we want to explore the possibilities of using human language-based models, LLMs, for model explainabilities. This survey provides a comprehensive overview of existing approaches regarding LLMs for XAI, and evaluation techniques for LLM-generated explanation,  discusses the corresponding challenges and limitations, and examines real-world applications. Finally, we discuss future directions by emphasizing the need for more interpretable, automated, user-centric, and multidisciplinary approaches for XAI via LLMs.
\end{abstract}





\maketitle

\section{Introduction}\label{sec:intro}
Recent advances in artificial intelligence (AI) have resulted in the development of more complex models, especially in the field of deep learning. AI models have demonstrated remarkable capabilities in various domains, from healthcare to finance \cite{khan2024leveraging}, \cite{cao2020ai}. However, as these AI models become more complex, it is challenging to understand how specific outputs are generated due to a lack of transparency \cite{arrieta2020explainable}. The lack of transparency in AI model decision-making is often referred to as a "black box" problem, which hinders trust and limits widespread adoption, particularly in critical fields such as healthcare and finance \cite{bhatt2020explainable}. Despite ongoing efforts to enhance the explainability of the AI model \cite{hosain2024explainable}, many experts without a background in machine learning still struggle to understand how these systems generate decisions. Enhancing transparency is therefore crucial to increasing trust and facilitating broader deployment across various domains. This makes it challenging for them to use these explanations to make better choices. For example, in healthcare, a doctor might not fully understand why a machine learning model recommends a particular treatment, making it hard for them to trust or act on the model's advice. In finance, a financial analyst may have difficulty interpreting how an AI system predicts market trends, which could lead to hesitation in relying on the model's predictions. Explainable AI (XAI) includes methods to improve the interpretability of neural networks and other state-of-the-art AI models, such as Convolutional Neural Networks (CNNs) for image recognition, Recurrent Neural Networks (RNNs) for sequence data, and Generative Adversarial Networks (GANs) for image generation, aiming to enhance transparency without compromising performance metrics, such as accuracy \cite{arrieta2020explainable}. \\
\indent XAI helps balance the tradeoff between making models understandable and maintaining their effectiveness. Explainability remains a significant challenge across various applications \cite{ribeiro2016trust}, \cite{xu2021learning}, \cite{camburu2018esnli}. By making model decisions understandable, XAI can build user trust, ensure accountability, and promote responsible and ethical use of these models. LLMs are becoming critical tools with broad applications across diverse domains. In medicine, they assist in tasks, such as diagnostics and personalized patient care \cite{thirunavukarasu2023large}. In finance, they support functions, such as risk assessment and market analysis \cite{wu2023bloomberggpt}. Additionally, in the field of natural language processing (NLP), LLMs are essential for tasks, such as text classification, summarization, and sentiment analysis. The extensive data on which LLMs are trained offers valuable potential for clarifying model architectures and the decision-making processes of AI models \cite{zhao2023explainability, amin2023chatgpt}.\\
\indent LLMs are an important link between complex AI models and XAI systems due to their natural language processing capabilities \cite{liu2023evaluating}. For example, instead of just making a prediction in a medical imaging AI model, the model might use an LLM to explain why it flagged a lung scan as abnormal, including highlighting patterns associated with specific diseases. LLMs help in XAI in many ways, such as understanding user questions to generate the appropriate explanation \cite{nguyen2023black}, \cite{shen2023convxai}, and \cite{slack2023talktomodel} also by directly explaining the complex ML model architecture and explanation to their output \cite{kroeger2023large}. For instance, \cite{bhattacharjee2024towards, memon2024llminformed} demonstrated the use of a simple prompting method to identify key features of an AI model's predictions and generate counterfactual explanations using LLMs. This approach highlights the potential of LLMs to improve understanding and transparency in AI decision-making processes, making AI systems more interpretable and trustworthy across diverse fields.\\
\indent Our survey discusses various approaches to LLM-based Explainability, as shown in Figure \ref{fig:taxonomy_explainability_llms} and in Table \ref{tab:overview_llm_xai}. We will specifically discuss three approaches: 1). The first approach, post-hoc explanations, corresponds to causal interpretability and focuses on analyzing why a specific input led to a particular output, providing explanations for output generated by the machine learning (ML) model; 2). The second approach, intrinsic explainability, corresponds with engineers' interpretability and involves designing machine learning model architectures using LLMs to make machine learning models more explainable.; 3). The third approach, human-centered narratives, corresponds to trust-inducing interpretability by enhancing explanations for the outputs generated by ML models through natural language, making the outputs more understandable and fostering user trust. For example, the AI model predicts a high probability that a patient develops hypertension in the next five years due to their history of high cholesterol, a family history of hypertension, and contributing factors, such as age and weight. While their blood pressure is normal, these risk factors suggest future concerns. This narrative helps the doctor understand the reasoning behind the prediction. This survey will also explore evaluation techniques for these explanations and how explainability can be applied in real-world applications. \\
\indent Then, we discuss the challenges and limitations of achieving explainability in AI models using LLMs, as illustrated in Figure \ref{fig:challenges}. We discuss this part in three topics: 1). understanding societal norms in terms of privacy; 2). managing the complexities of AI systems;  3). adapting LLMs to domain-specific tasks. We examine how various state-of-the-art LLM architectures emphasize different aspects of explainability using saliency maps, as shown in Figure \ref{fig:saliency}. Finally, we outline future directions for improving explainability through LLMs. By integrating insights from both model architecture and narrative approaches, our survey aims to present a comprehensive understanding of how LLMs can improve explainability in AI systems.
\section{Background}
The urgent need to interpret and trust AI systems has increased manifold in recent years. The evolution of XAI would help with this urgent need by making AI decisions transparent and easy for human beings to comprehend. Meanwhile, the tremendous advancements in LLMs have opened up new possibilities of generating detailed and dynamic explanations for AI systems. Additionally, LLMs provide experience in ML and AI to nontechnical and technical people for better understanding and effective acting on model decisions, opening up AI technologies to larger users. Merging XAI techniques with the capabilities of LLMs enables researchers and practitioners to create more interpretable AI across domains that are reliable, fair, and user-friendly. The next sections discuss XAI, explore how LLMs contribute to explainability, and identify areas for future research in AI and NLP.
\begin{figure*}[htbp]
    \centering
    \includegraphics[width=\linewidth]{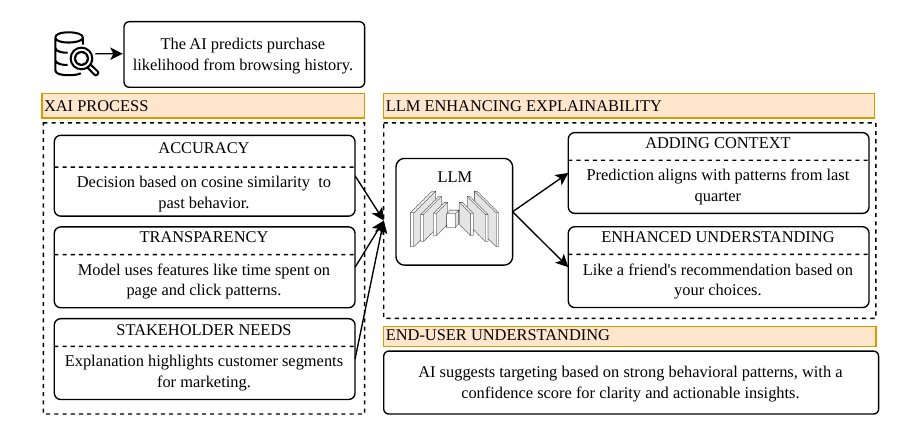}
    \caption{Explainable AI and Its Intersection with LLMs}
    \label{fig:xai_llms}
\end{figure*}
\subsection{Explainable AI}
Explainable AI (XAI) refers to a range of methods that make AI model outputs more transparent and understandable to humans, providing access to the inner workings of AI models so that users can understand the design and logic behind AI systems. XAI creates models that are more understandable, trustworthy, and ethically aligned with user expectations and regulatory requirements. XAI helps increase the interpretability of complex models by clarifying the most influential factors of AI predictions, such as the key input features identified in machine learning models \cite{ribeiro2016trust}.\\
\indent XAI improves transparency, fairness, and reliability, to seamlessly align with the ethical principles. Regulatory bodies, such as the European General Data Protection Regulation (GDPR) emphasize transparency by encouraging AI systems to be understandable and accountable to both users and regulatory bodies \cite{bostrom2018ethics}. In fairness, XAI helps developers identify biases and limitations within models \cite{pahde2023reveal}, allowing software developers to debug algorithms and improve their decision-making processes. For reliability, industry professionals, such as product engineers, use XAI-driven insights to responsibly integrate AI solutions into applications \cite{weld2019challenge}, ensuring that products are reliable. In short, XAI offers clear explanations that help end-users understand and trust AI outputs.
\subsection{Leveraging Large Language Models for AI Model Explainability:}\label{sec:leverage_llm}
Large Language Models (LLMs) have developed significantly, starting with the basic models, such as GPT and BERT, and then advancing to more complex and larger models, such as LLaMa \cite{touvron2023llama2}. Transformer architectures in LLMs enable LLMs to process large amounts of data efficiently. With billions of parameters, LLMs can highly capture complex and important features in the input \cite{touvron2023llama2}. LLMs are trained on massive datasets \cite{penedo2023refinedweb} and fine-tuned to enhance their ability to follow instructions \cite{ouyang2022training}. LLMs can now tackle a wide range of tasks beyond text generation, providing valuable insights for model explainability. Traditional methods, such as Principal Component Analysis (PCA) rely on statistical techniques for feature selection. In contrast, large language models (LLMs) utilize extensive contextual knowledge to identify and emphasize important features based on input data dynamically. The adaptability of LLMs to dynamically focus on important features allows LLMs to offer context-sensitive explanations, making them particularly useful in complex domains.\\
\indent For example, in the communication domain, where during communication, understanding the contributing factors, such as signal interference, propagation loss, multipath fading, noise, and weather conditions, is a challenge \cite{mohsin2025hierarchical, mohsin2025vision, bhattacharya2025task}. LLMs help clarify the underlying complexities. By breaking down complex models and filling knowledge gaps, LLMs provide valuable insights into dynamic environments, demonstrating their potential to enhance AI model explainability in various fields, such as communication and cybersecurity \cite{zou2024genainet, erak2024leveraging} \cite{feng2023knowledge}.\\
\indent In the realm of cybersecurity, LLMs have proven effective for explainable AI. For example, LLMs generate human-readable insights to identify the detection patterns on very specific malware or phishing attack detection, network anomalies, and so on. LLMs improve the transparency of an Intrusion Detection System (IDS)'s decision-making process by helping users understand why specific alerts were triggered. For example, LLMs can provide explanations about patterns in network traffic, highlight anomalous behaviors, or detail the specific rules or thresholds that caused an alert \cite{elouardi2024survey}. LLMs improve explainability as shown in Figure \ref{fig:xai_llms} and build trust by making the reasoning behind alerts and actions more accessible to users \cite{cyber2024}. LLMs, therefore, generate explanations and facilitate the creation of more intuitive and interpretable AI systems.
\begin{figure*}[h]
    \centering

    \includegraphics[width=\linewidth]{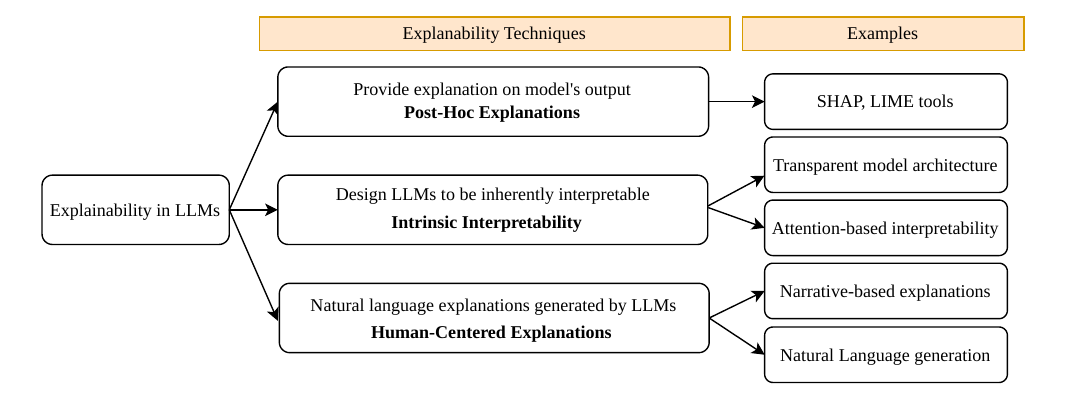} 
    \caption{Techniques for explainability in Large Language Models (LLMs). The diagram identifies three broad categories of approaches to explaining: Post-hoc Explanation, Intrinsic Interpretability, and Human-Centered Explanations, providing examples of methods in each category.}
    \label{fig:taxonomy_explainability_llms}
\end{figure*}
\section{Overview of Explainability Techniques} \label{sec:techniques_section}
Techniques for explainability in AI systems using Large Language Models (LLMs) are categorized into 1) post-hoc explanations, 2) intrinsic interpretability, and 3) human-centered explanations \cite{zhang2022developing}. Post-hoc explanations, such as SHAP and LIME, clarify model predictions by analyzing feature contributions after decisions are made. Intrinsic interpretability focuses on designing inherently explainable models, such as Chain of Thought (CoT) reasoning, which breaks tasks into logical steps. Human-centered approaches incorporate user feedback to ensure explanations are clear and tailored to human needs, enhancing the overall interpretability of AI models. A visual representation of explainability techniques is provided in Figure \ref{fig:taxonomy_explainability_llms}. 
\subsection{Post-hoc Explanations}\label{sec:post_hoc}
Post-hoc explanations clarify the decisions that large language models make after generating predictions. Post-hoc explanations help users to understand the reasoning behind a model's decision by analyzing how the model arrived at a specific conclusion. The post-hoc technique provides two types of explanations \cite{SETZU2021103457}: 1) global explanations and 2) local explanations. Global explanations illustrate the overall behavior of the AI model, and local explanations focus on a single prediction and highlight the important input features that contributed to that result.\\ 
\indent For example, consider an email classification model that predicts whether an email is spam. A global explanation identifies features, such as keywords (e.g., "free," "winner," "click here") and patterns (e.g., excessive exclamation marks or links) that influence the model's spam predictions across all emails. In contrast, a local explanation for a specific email focuses on the terms or phrases that contributed most to that particular prediction, such as "claim your prize," "congratulations," or a high count of hyperlinks, providing insight into why the model classified that email as spam. \\
\indent Post-hoc explanation tools interpret and explain a model’s decisions after the model makes a prediction. The methods and tools for post-hoc explanations include Integrated Gradients (IG), Local Interpretable Model-Agnostic Explanations (LIME), and Shapley Additive explanations (SHAP). Integrated Gradients (IG) \cite{ig_xai} explains individual predictions by identifying the most effective parts of the input concerning the output. IG reveals the features that have the most impact on specific predictions by analyzing how changes in the input affect the output. \\
\indent LIME focuses on local explanations by identifying the key features of a single prediction. LIME modifies the version of the dataset around the prediction and, analyzes the feature importance using techniques, such as LASSO or principal component analysis (PCA). In contrast, SHAP provides both local and global explanations. SHAP explains individual predictions by using Shapley values (a concept from game theory) to measure the importance of each feature \cite{park2021fast}.\\
\indent In local explanations, SHAP works by analyzing changes in the model prediction by removing different features and determining the contribution of every feature to the prediction. The Shapley values from SHAP provide the relation between the specific prediction and the model's behavior. For example, in a spam email detection model, SHAP shows that the presence of the word "free" increases the likelihood of the email being spam, while the absence of words, such as "meeting" reduces the likelihood and helps to explain the model that specifies email as spam \cite{Bouke2024}. SHAP extends its utility beyond local explanations by aggregating feature contributions across the entire dataset to provide a global explanation. For example, in the same spam email detection model, a global explanation reveals that certain features, such as frequent occurrences of promotional words ("offer," "discount"), and patterns, such as high density of hyperlinks, are consistently influential in the model's spam classifications across all data points.\\
\indent Post-hoc explanations enhance model transparency by analyzing factors that influenced the output, such as specific input (local explanation) or general trends across the data (global explanation). The increase in transparency reduces the perception of the model as a "black box" and reassures users that the model is making decisions based on understandable and reasonable criteria, rather than arbitrary or hidden processes.\\
\indent However, post-hoc methods have limitations. The explanations can be unstable, as small changes in the input data can result in inconsistent results. For example, a slight variation in the input features might significantly alter the explanation, even though the model's prediction remains the same. Furthermore, post-hoc explanations focus on interpreting the model's predictions rather than fully capturing model's internal workings. For instance, in a spam email detection model, a post-hoc method might explain, why the model predicts spam by identifying relevant keywords or patterns, but may not reveal the detailed processes or computations the model used to make that decision. 
\subsection{Intrinsic Interpretability}\label{sec:intrinsic}
XAI has two main approaches: explaining predictions and designing interpretable model architectures. Post-hoc explanations focus on clarifying the model's decisions after the model generates the prediction (the first approach), and intrinsic interpretability focuses on building models that are understandable by design (the second approach). The intrinsic interpretability approach involves structuring the model architecture using large language models (LLMs), to make the model's behavior inherently explainable. For instance, in a sentiment analysis task, an LLM equipped with attention mechanisms can show phrases, such as "excellent service" or "poor quality" that directly contribute to a positive or negative sentiment classification. Attention weights act as built-in explanations, making the model’s decision-making process easier. With the help of built-in explanations, the model's decisions are understandable by design, without needing external explanation tools \cite{tull2024compositionalinterpretabilityxai}. \\
\indent A notable technique in this category is Chain of Thought (CoT) reasoning \cite{chu2023survey}, which breaks down complex problems into smaller, manageable steps, allowing models to follow a logical progression toward conclusions. The logical flow reduces error propagation and makes the model’s logic easier to follow. For example, in predicting a patient's condition, CoT reasoning divides the decision-making process into sequential steps: first identifying symptoms, then evaluating potential diagnoses, and finally analyzing the relationships between symptoms and diagnoses. The sequential steps help to make the overall reasoning of AI model more interpretable and clear. \\
\indent Methods in the Chain of Thought (CoT) include 1) Guided CoT Templates and 2) ReAct (Reasoning and Acting). Guided CoT templates provide logical steps to guide the model's reasoning process and ReAct incorporates task-specific actions to enhance the intrinsic interpretability of models \cite{wei2023chainofthoughtpromptingelicitsreasoning}. Guided CoT templates provide structured, predefined templates that help models follow a logical, multistep reasoning process. For example, in a math problem-solving task, a Guided CoT template prompts the model to break the problem into smaller steps (e.g., “First, calculate the sum, then subtract the result from the total”), to make the model’s internal reasoning more understandable, and improve the intrinsic interpretability by revealing how the model arrives at each step. ReAct enables models to perform iterative task-specific actions with step-by-step reasoning. For instance, in a navigation task, ReAct guides the model to reason about the best route, take a step (move forward), and then reason again based on the new position.  The iterative reasoning and action increase the interpretability of AI model by real-time tracing of the decision-making process, effectively correcting the model’s explanation and revealing the internal actions and decisions of the model during execution. For example, in a question-answering model, ReAct involves the following steps: 1) prompting the model to read a question, 2) extracting relevant information from the text, 3) reasoning about the correct answer, and 4) providing an explanation for its choice.\\
\indent Intrinsic interpretability helps in architecture-based XAI by following a structured process. The first process involves LLMs generating explanations, training deep learning models on these explanations, and modifying the model architecture accordingly. Refining the architecture improves the accuracy and clarity of the generated explanation. The step proves particularly useful for tasks that demand both reasoning and interaction with external systems or calculations to ensure the necessary explanations are clear and precise. \\
\indent After generating the explanation, the second process checks the reliability of the explanation. Factual accuracy and logical correctness are two key aspects of a reliable architecture-based XAI. \cite{ccyang2023survey}. Factual accuracy ensures that the explanation aligns with real-world information and provides the correct output based on the model's inputs. Logical correctness, on the other hand, focuses on the reasoning process, ensuring that the steps taken to conclude follow a coherent and consistent pattern. Factual accuracy alone does not guarantee reliable explanations; even incorrect outputs can be presented in convincing ways. For example, a model may explain a medical disease prediction, such as identifying diabetes risk based on patient symptoms, that is accurate but based on incorrect assumptions or unrelated data, such as a spurious correlation between age and unrelated dietary habits.  Studies \cite{golovneva2022roscoe} show that factual accuracy combined with logical consistency leads to more reliable explanations, and helps build user confidence in the model’s outputs. For example, in legal case analysis, LLMs generate an explanation that accurately cites relevant laws (factual accuracy) while also outlining a logical step-by-step reasoning process to apply relevant laws to the case. The combination ensures that the explanation is both correct and easy for users to trust and understand." The above discussion shows that Leveraging the LLMs for model generation, emphasizing both factual accuracy and logical correctness, and combining clear reasoning with interpretable outputs improve the architectural reliability of XAI.\\
\begin{table*}[htbp]
\centering
\caption{Overview of XAI with LLMs}
\label{tab:overview_llm_xai}
\begin{tabular}{p{4cm} p{5cm} p{4cm}}
\toprule
\textbf{Topic} & \textbf{Key Points} & \textbf{Sources} \\
\midrule
Background & XAI promotes transparency, trust, and regulatory compliance & \cite{bostrom2018ethics}, \cite{weld2019challenge}, \cite{ribeiro2016trust}, \cite{touvron2023llama2}, \cite{penedo2023refinedweb}, \cite{zou2024genainet} \\
\midrule
Post-hoc explanations & LIME for local interpretations, SHAP for global and local insights & \cite{shap}, \cite{park2021fast}, \cite{SETZU2021103457}, \cite{ig_xai}, \cite{Bouke2024} \\
\midrule
Intrinsic interpretability & Chain of thought reasoning, guided templates for logical progression & \cite{chu2023survey}, \cite{wang2024llm}, \cite{golovneva2022roscoe}, \cite{lanham2023measuring}, \cite{yeo2023interpretable}, \cite{tull2024compositionalinterpretabilityxai} \\
\midrule
Human-centered explanations & User-focused design, collaborative feedback methods & \cite{eiband2018bringing}, \cite{amershi2019guidelines}, \cite{neerincx2019socio}, \cite{ehsan2020human}, \cite{wolf2019explainability}, \cite{JIANG2024100078} \\
\midrule
Evaluation of explanation  & Qualitative and quantitative evaluation of LLM generated explanation & \cite{zhao2023explainability}, \cite{cao2023learn}, \cite{nauta2023anecdotal}, \cite{dong2017improving}, \cite{chen2021towards}, \cite{sokol2024does}, \cite{jacovi2020towards}, \cite{burton2023natural}, \cite{min2023factscore}, \cite{ichmoukhamedov2024good} \\
\midrule
Benchmark datasets &  Such as e-SNLI, CoS-E, and ECQA help in training the LLMs to generate clear, human-understandable explanations & \cite{camburu2018snli},\cite{rajani2019explain}, \cite{aggarwal2021explanations}, \cite{jansen2018worldtree}, \cite{mihaylov2018can}, \cite{chen2024xplainllm}, \cite{friel2024ragbench}, \cite{mathew2021hatexplain}. \\
\midrule
Application of XAI in different domains & Application of explainability using LLMs & \cite{umerenkov2023decipheringdiagnoseslargelanguage}, \cite{tropicalmed9090216}, \cite{dave2024integratingllmsexplainablefault}, \cite{ehsan2020human}, \cite{umerenkov2023deciphering}, \cite{yang2023towards}, \cite{benary2023leveraging}, \cite{luz2024enhancing}, \cite{feng2023empowering}, \cite{thanh2024krag} \\
\midrule
Challenges in XAI & Sensitive data, societal diversity, complex multi-source and other factors & \cite{huang2023bias}, \cite{caliskan2023artificial}, \cite{Klein2020}, \cite{recommend_biases}, \cite{82ac6ae737e94dd6809b8d17b2621a68}, \cite{hua2024up5unbiasedfoundationmodel} \\
\midrule
Analysis of Feature importance & Feature importance using saliency map across LLMs & \cite{Zhou2021}, \cite{mitllm}, \cite{jiang2023mistral}, \cite{Bai2023}, \cite{openai}, \cite{koo2024benchmarking}, \cite{touvron2023llama2}, \cite{almazrouei2023falcon40b}, \cite{jiang2023mistral}, \cite{groeneveld2024olmo} \\
\midrule
Future directions & Human-in-loop automation, visual explanations, interdisciplinary approach & \cite{do2024facilitatinghumanllmcollaborationfactuality}, \cite{zhang2024explainable}, \cite{Selvaraju_2019}, \cite{ehsan2020human}, \cite{SCHOONDERWOERD2021102684}, \cite{Cabour2023}, \cite{know} \\
\bottomrule
\end{tabular}

\end{table*}
\subsection{Human-Centered Explanations}\label{sec:human_centre}
Both post-hoc and intrinsic interpretability methods significantly improve XAI by cross-checking with factual ground truth. However, the challenge arises in presenting this explanation as understandable to the end user. Human-centered  approach that designs XAI with the end user in the workflow\cite{eiband2018bringing, amershi2019guidelines} addresses the challenge of comprehensible XAI. The approch provides guidelines for researchers and developers to create explanations that cater to human needs. It also discusses how to apply human-centered design methods to better understand the social and technical aspects for better human-AI interaction \cite{neerincx2019socio, ehsan2020human}. A human-centered approach to XAI involves determining what, when, and how to explain information to users. Interviews, focus groups, and surveys are common ways to include users in the workflow of the model development process. \\
\indent In previous research, the approach of keeping the end user in the workflow plays an important role in improving the transparency of AI models. For example, research \cite{eiband2018bringing} outlines ways to enhance transparency in intelligent systems by involving users in a step-by-step design process for developing explanations. Research \cite{wolf2019explainability} illustrates the importance of using user scenarios early in the design process to identify users' needs for explanations and to guide further development.\\
\indent Collaboration of end user with AI model, helps developers understand user values and the social context of human-AI interactions. The human-centered technique uses contexts of human-AI interaction to refine and improve the explanation. Specifically, the technique captures user preferences, comprehension levels, and relevant contextual factors to refine the explanation. The refined explanations are then tested with users in realistic work settings, focusing on important information for clear comprehensibility between user and AI model \cite{JIANG2024100078}. For example, the DoReMi framework applies the four Human-centralized design principles (understand, define, design, evaluate) to XAI, enhancing existing methods by generating knowledge through design patterns that connect user requirements with design solutions.\\
\indent The researcher \cite{martens2023tell} generates human-centered explanations using counterfactual (CF) information in combination with large language models (LLMs). The study uses various prompts for the LLMs by giving model prediction as the context to generate a narrative based on the prompts. For example, if a model predicts a low credit score, a prompt might ask the LLMs to explain how a higher income or lower debt could change the prediction. Depending on users' reactions to the generated explanation, different adjustments are made for efficient XAI. In short, a human-centered approach focuses on gathering essential information during model development and then providing explanations through LLMs while keeping humans in the workflow using the appropriate prompts. 
\section{Evaluating LLM-Generated Explanations}
In the previous section \ref{sec:techniques_section}, the survey discussed several explanation techniques using LLMs. However, evaluating the aforementioned explanation techniques to capture the model's reasoning processes remains a significant challenge \cite{zhao2023explainability}. Evaluation can be divided into two categories: 1) qualitative and 2) quantitative evaluation as shown in Table \ref{tab:evaluation-metrics}. Qualitative evaluation focuses on how easy the explanations are to understand and read, emphasizing aspects, such as 1) comprehensibility and human understanding, and 2) controllability. On the other hand, quantitative evaluation, looks at things, such as relevant, and efficient explanations, such as 1) faithfulness and 2) plausibility.\\
\indent \textbf{Comprehensibility and human understanding} focus on the ease with which the explanation can be understood and the clarity with which it conveys the model's reasoning to humans \cite{Lim2019WhyTE, sokol2024does}.  For example, if an LLM explains the prediction by highlighting certain words in the sentence, the explanation should be easy for a human to follow. For example, in the case of sentiment analysis, a comprehensible explanation would read something, such as: "Review was classified as positive mainly for three key phrases: 'masterful direction' indicates the high quality of filmmaking, 'compelling performances' indicate that the acting was great, and 'innovative storytelling' indicates creative merit. Such phrases occur frequently in positive movie reviews within our training data. The explanation is comprehensible because it breaks down the model's reasoning into distinct components, links each to the corresponding evidence, and connects distinct components to commonly understood concepts in film criticism.\\
\indent \textbf{Controllability} is another important qualitative evaluation aspect that refers to the interactivity and adjustability of explanations \cite{cao2023learn}. For example, enabling a user to set the focus of an explanation, such as highlighting select parts of the model's decision process that are interesting or of interest to users, would increase the controllability of the explanation. In LLM-generated explanations, controllability allows users to provide feedback to improve the explanation. For example, a user identifies the unclear parts of the model and ask the LLM to improve its response. One way to test how well this works is through user studies, where participants give feedback on explanations generated by the model, such as highlighting sections that don’t make sense or aren’t relevant. Based on this feedback, the model’s ability to adjust its explanations and how satisfied users are with the updated versions would be used to measure its flexibility/controllability. A few works study this process iteratively, providing feedback and improving explanation quality by measuring the effects of human input on explanations \cite{nauta2023anecdotal, dong2017improving}.\\ 
\indent Some researchers argue that explanations that look understandable to humans might not be accurate or true to the model's actual logic \cite{jacovi2020towards}. For example, an explanation might seem accurate but could still be wrong because the model was trained on flawed data, leading to incorrect relationships \cite{nauta2023anecdotal}. Researchers stress the importance of using more objective, quantitative methods to evaluate explanations to address the issue of misleading explanations. For example, checking whether an explanation is truly faithful to the model, rather than just looking reasonable, can help prevent mistakes. Research \cite{burton2023natural} focuses on the quantitative evaluation of explanations generated by large language models (LLM), by assessing aspects: 1) faithfulness and 2) plausibility. \\
\indent \textbf{Faithfulness} is the degree of accuracy with which an explanation represents the model's actual decision-making process \cite{kadir2023evaluation}. A faithful explanation should align with at least the following: 1) internal logic and cause-and-effect relationship established by the model; 2) how the model has been designed and the data the model has been trained on; and 3) whether the decision of the model is reproducible using features represented in the explanation. A faithful explanation in a medical diagnosis system, for instance, clearly states the features in the patient that help in model decisions, such as his or her age, symptoms, test results, previous case history, and the likes. A faithful explanation shows the transparent cause-and-effect relationships regarding how particular symptoms or positive test results add to the diagnosis. It should be consistent with what the model has learned; that is, it uses the same patterns as picked up during training, such as known associations between symptoms and conditions. Confidence scores or weights on each factor are meant to give an idea about the influence each feature in the decision-making process resulted in the diagnosis and further to demonstrate that the given decision can be reproduced using these explained features \cite{burton2023natural}.\\
\indent \textbf{Plausibility}, by contrast, evaluates the logical coherence and domain consistency of explanations \cite{nauta2023anecdotal}. The plausible involves 1) aligning explanations with domain knowledge, 2) ensuring logical consistency, 3) verifying causal relationships, and avoiding contradictions or impossible claims. For instance, a plausible explanation from a climate prediction model would be based on established meteorological principles and valid cause-effect relationships \cite{jacovi2020towards}. The difference between faithfulness and plausibility is important: an explanation can be plausible, it makes logical sense to the human not faithful, and it does not represent the model's internal reasoning. For example, a cat prediction model might predict the presence of a cat in an image because it detected whiskers and a tail. However, the prediction may be based on unrelated features, such as texture patterns.
\begin{table*}[t]
\caption{Comprehensive Evaluation for LLM Explanations}
\label{tab:evaluation-metrics}
\centering
\begin{tabular}{p{0.15\linewidth}|p{0.15\linewidth}|p{0.35\linewidth}|p{0.25\linewidth}}
\hline
\textbf{Category} & \textbf{Metric} & \textbf{Description} & \textbf{Example} \\
\hline
\multirow{2}{*}{Qualitative} 
& Comprehensibility  and human understanding \cite{Lim2019WhyTE, sokol2024does} & Ease of understanding and clarity in delivering the model's reasoning to humans & Explaining sentiment analysis by highlighting key phrases, such as "masterful direction" and "innovative storytelling" \\
\cline{2-4}
& Controllability \cite{cao2023learn, chen2021towards} & Interactivity and adjustability of explanations, enabling users to provide feedback to refine explanations & Users highlighting unclear parts of the explanation for improvement \\
\hline
\multirow{2}{*}{Quantitative} 
& Faithfulness \cite{burton2023natural} & Accuracy in representing the model's internal decision-making process, including causal relationships and alignment with training data & A medical diagnosis system showing specific patient features and their weights that informed the decision \\
\cline{2-4}
& Plausibility \cite{jacovi2020towards, nauta2023anecdotal} & Logical coherence and domain consistency, ensuring alignment with established knowledge & A climate prediction model providing explanations consistent with meteorological principles \\
\hline
\end{tabular}
\end{table*}
\section{Benchmark Datasets for Explainable AI with LLMs}
Researchers need special data to test these explanation generation features of LLMs and make them better \cite{zytek2024explingo}. Such datasets can be used for training LLMs to explain AI model's decision by providing structured sentences \cite{liu2024datasets}. The goal is to generate an explanation that is clear, transparent, and easy to understand for humans. Benchmark datasets include examples, such as human-written explanations, steps of reasoning of prediction, and structured knowledge that aligns with factual information  \cite{chen2024xplainllm}. These examples enable AI systems to become skilled in explaining their thought processes.\\
\indent Benchmark datasets are important because they give researchers a standard way to train and measure the quality of explanations produced by LLMs. For example, datasets allow researchers to measure whether the explanations are accurate, complete, easy to understand, and consistent across similar situations. By studying these datasets, researchers can better understand how to make AI systems that are not only clear but also more transparent and user-friendly. Each dataset has been designed to focus on specific aspects of explanation, such as clear reasoning, scientific knowledge (detailed annotations and structured reasoning examples), or ethical decision-making (containing scenarios with diverse ethical considerations).As shown in Table \ref{tab:dataset_overview}, the datasets include 1) e-SNLI (Enhanced Stanford Natural Language Inference) \cite{camburu2018snli}, 2) CoS-E (Common Sense Explanations) \cite{rajani2019explain}, 3) ECQA (Explanation CommonsenseQA) \cite{aggarwal2021explanations}, 4) WorldTree \cite{jansen2018worldtree}, 5) OpenBookQA with Explanations \cite{mihaylov2018can}, 6) XplainLLM \cite{chen2024xplainllm}, 7) RAGBench \cite{friel2024ragbench}, and 8) HateXplain \cite{mathew2021hatexplain}. The overview of each dataset is given below.\\
\indent First we have the \textbf{e-SNLI (Enhanced Stanford Natural Language Inference)} dataset \cite{camburu2018snli}, which is an extension of the Stanford Natural Language Inference dataset that includes sentence pairs annotated with entailment labels (e.g., "entailment," "contradiction," or "neutral") and human written explanations. For example, if the first statement is "A dog is running through the park" and the second statement is "An animal is outside," the two statements match, then the label is "entailment," and the explanation is "A dog is a type of animal, and running through a park implies being outside." The second dataset is the \textbf{CoS-E (Common Sense Explanations)} dataset \cite{rajani2019explain}, which provides the multichoice common-sense questions with steps with human written explanations, to enable the model to generate everyday reasoning. For example, a question in the dataset might ask, "Why would someone wear sunglasses?" with multiple-choice answers, such as "To see better at night," "To block the sun," or "To look fashionable." The correct answer, "To block the sun," is accompanied by a detailed explanation: "Sunglasses are designed to protect the eyes from bright sunlight." The third dataset is \textbf{ECQA (Explanation CommonsenseQA)} dataset \cite{aggarwal2021explanations}, which extends the CommonsenseQA dataset \cite{talmor2018commonsenseqa} by adding the detailed explanation for the answers to the commonsense reasoning questions. For example, a question, such as "Why would someone lock their car?" with answer choices such as "1) To prevent theft," "2) To keep it clean," and "3) To make it lighter" is combined with the correct answer, "To prevent theft." ECQA provides detailed explanations with positive reasons, such as "Locking the car secures the doors and prevents unauthorized access" and negative reasons, such as "Leaving the car unlocked makes it easier for thieves to steal belongings." These reasons help models identify the correct answer, explain it clearly, and prevent incorrect options, to improve the reasoning and explanatory abilities of models. Additionally, \textbf{WorldTree} dataset \cite{jansen2018worldtree}, organizes scientific reasoning questions into structured explanation graphs to train the large language models for step-by-step scientific reasoning. For example, for the question, "Why does a metal spoon feel hotter than a plastic spoon in hot water?" with the correct answer, "Because metal is a better conductor of heat," WorldTree dataset provides an explanation graph linking facts like "Metal conducts heat efficiently," "Conductors transfer heat more effectively," and "Plastic is a poor conductor." These explanation graphs combine domain knowledge and logical steps to help the models articulate clear and scientific reasoning processes. Similarly, \textbf{OpenBookQA} \cite{mihaylov2018can} dataset improves elementary science questions with the help of knowledge-based explanations. For example, the question "Which of these would let the most heat travel through?" the correct answer is "a steel spoon in a cafeteria". The answer is explained using the scientific fact "Metal is a thermal conductor" and the common knowledge "Steel is made of metal" and "Heat travels through a thermal conductor." Training the LLMs on this openbookqa dataset can enhance the ability to explain the relation between the science questions and answers. \textbf{XplainLLM} dataset \cite{chen2024xplainllm} combine the question-answers with the knowledge graphs in the form of triplets. These triplets ensure transparency by using factual grounded reasoning in the knowledge graphs. For example, for the question "The people danced to the music, what was the music like for them?" with options such as "soothing," "vocal or instrumental," "loud," "universal," and "enjoyable," the correct answer is "enjoyable." XplainLLM provides explanations grounded in knowledge graphs, detailing why "enjoyable" is the correct choice while justifying why the other options are less suitable. Training the LLM on this dataset enhances transparency and reliability by linking the LLM's reasoning to structured and factual knowledge. The other dataset is the \textbf{RAGBench} \cite{friel2024ragbench}, which is a domain-specific dataset for using Retrieval-Augmented Generation (RAG) systems with a focus on explainability. Each sample in the RAGBench includes a query, the retrieved evidence, and the generated response using a query. For instance, in a technical support domain, a query such as "How do I reset my device?" would be paired with relevant text from a user manual with the details of the reset process. This structure allows RAG systems to generate accurate and contextually accurate responses, with the retrieved evidence providing transparency into the model's decision-making process. Training on RAGBench can enhance the ability of the model to produce explainable and reliable outputs in domain-specific applications. The last dataset is the \textbf{HateXplain} dataset \cite{mathew2021hatexplain} provides explanation in the hate speech detection with annotations for an unbiased generation.  Each sample of the HateXplain dataset includes hate speech classification and a rationale explaining the classification. For example, a post labeled as "hatespeech" might target a specific community, with annotators highlighting specific words or phrases that justify this classification. The detailed annotation aids in understanding the model's decision-making process and helps in reducing unintended bias towards target communities.
\begin{table*}[ht]
\centering
\caption{Overview of Datasets for LLM-Based Explainability}
\label{tab:dataset_overview}
\begin{tabular}{p{3cm} p{6cm} p{6cm}}
\toprule
\textbf{Dataset} & \textbf{Focus/Use} & \textbf{Example} \\
\midrule
\textbf{e-SNLI} \cite{camburu2018snli}  & Human-written explanations for tasks to find logical relation. & Sentence and label pairs, such as "entailment," "contradiction," or "neutral," e.g., "A dog is a type of animal, and running through a park outside." \\
\textbf{CoS-E} \cite{rajani2019explain} & Multichoice common-sense reasoning with explanations. & Example: "Why wear sunglasses?", pair with the answer: "To block the sun," with explanation: "Sunglasses protect eyes from bright sunlight." \\
\textbf{ECQA} \cite{aggarwal2021explanations} & Commonsense reasoning with detailed explanations. & "Why lock a car?" with answer: "To prevent theft," with explanation: "Locks secure doors, preventing unauthorized access." \\
\textbf{WorldTree} \cite{jansen2018worldtree} & Scientific reasoning using structured explanation graphs. & Example: "Why does a metal spoon feel hotter than plastic in hot water?" with the answer: "Metal conducts heat efficiently," and explanation: Graph with domain facts, such as "Metal is a conductor." \\
\textbf{OpenBookQA} \cite{mihaylov2018can} & Elementary science questions with knowledge-based explanations. & Example: "Which lets heat travel best?" with answer: "Steel spoon," and explanation: "Metal is a thermal conductor, and steel is made of metal." \\
\textbf{XplainLLM} \cite{chen2024xplainllm} & Factual reasoning with knowledge graphs in triplet form. & Example: "Why did people dance to music?" with answer: "Enjoyable," and explanation: Knowledge graph linking "enjoyable" with relevant reasoning. \\
\textbf{RAGBench} \cite{friel2024ragbench} & Retrieval-Augmented Generation with explainability for specific domains. & Example: "How to reset a device?" with explanation: Retrieved evidence from a user manual supports the generated response. \\
\textbf{HateXplain} \cite{mathew2021hatexplain} & Hate speech detection with unbiased explanations. & Example: A flagged post with highlighted words or phrases explaining the "hate speech" classification. \\
\bottomrule
\end{tabular}
\end{table*}
\section{Real-World Applications of LLM Explainability}
LLMs are playing a vital role in enhancing explainability and confidence in AI-driven decisions across a wide range of domains, ranging from healthcare to finance. Applications in healthcare, for instance, include the use of LLM in explainability to trust AI-driven medical decisions and predictions \cite{tropicalmed9090216, dave2024integratingllmsexplainablefault, umerenkov2023decipheringdiagnoseslargelanguage}. Suppose, for instance, that a doctor makes a diagnosis using an AI system. An LLM can break down the relationship between fever and chronic cough that leads to pneumonia, enable the doctor to consider the AI's explanation and make treatment decisions based on it. In finance, LLMs contribute to explainability in complex decision-making processes, such as in credit risk assessment or fraud detection. For example, if an AI system flags a transaction as potentially fraudulent, an LLM can give a detailed breakdown of the various factors influencing this decision, such as unusual amounts of transactions, atypical locations, or patterns inconsistent with the user's historical behavior \cite{feng2023empowering}. In this way, explainability allows financial institutions to review the flagged cases thoroughly and ensures that decisions are made according to regulatory standards and customer trust.\\
\indent In \textbf{healthcare}, LLMs improve the process of drug discovery \cite{zheng2024large, zhavoronkov2019deep} by explaining the interactions of molecules at chemical and biological levels \cite{liu2024large}. For example, in antibiotic development, LLMs analyze how a drug molecule binds to bacterial enzymes, and explain why certain compounds effectively kill bacteria while others do not. The explanation generated by LLMs ensures that scientists understand the combination of specific molecules to develop a drug and its impact on the human body. In drug discovery, LLMs help identify the molecule that works best against a disease \cite{blanchard2022language}. For example, in the design of new antiviral medications, LLMs process vast biomedical datasets to determine which molecules can bind to a virus's proteins and, therefore, block its ability to replicate. By breaking down these interactions step by step, LLMs help select the most promising drug candidates for further testing. In addition to the design of new antiviral medications, LLM-generated explanations also help predict the effects of drugs on the human body \cite{chaves2024tx}. For instance, if a new drug for heart disease is developed, an LLM can explain how its molecular structure will interact with heart cells and whether it may cause side effects such as high blood pressure. This transparency allows researchers to make informed decisions early in the drug development process, saving time and resources. \\
\indent In \textbf{finance}, explainability builds transparency into credit scoring systems and fraud detection \cite{luz2024enhancing, feng2023empowering}. An LLM might point out the load application refusal and potential factors behind it, such as a low credit score, high debt-to-income ratio, or missed payments. Such transparency not only increases trust among customers but also aids users in taking correct measures during loan applications. Similarly, in predicting currency trends, LLMs improve explainability in AI models by sentiment analysis of news articles and reports. LLMs analyze and explain particular sentiments, such as general optimism about the performance of an economy or concern over geopolitical instability, that influence the prediction of currency trends. Then, the  historical price data is combined with trading patterns for clear and accurate forecasts of the currency movement; therefore, financial analysts use it to make better-informed decisions \cite{limonad2024monetizing}.\\
\indent In \textbf{legal applications}, LLMs enhance the interpretability of AI-driven legal analysis and document review systems \cite{billi2023large, thanh2024krag}. When analyzing contracts or legal precedents, LLMs can clearly express the reasoning behind flagging specific clauses as potentially problematic or explain the reason behind the relevancy of certain cases to current litigation. For example, during a contract review by the lawyer, an LLM can explain why a particular compensation clause is risky by underlining that it is too broad compared to industry norms. Explainability provides lawyers with better decision-making and raises confidence in AI-assisted legal research.\\
\indent \textbf{Manufacturing and industrial applications} benefit significantly from LLM-enhanced explainability in predictive maintenance systems \cite{trivedi2024explainable, garcia2024framework}. For example, suppose an AI system predicts an imminent failure of a machine. In that case, the LLM can explain how certain vibration patterns, temperature readings, and other operational data contribute to this prediction, and enable the maintenance team to make informed decisions about preventive actions.\\
\indent In \textbf{educational technology}, Explainability using LLMs help in personalized learning recommendation.  \cite{laak2024ai, ng2024educational}. For example, when a high school student struggles with algebra, LLMs such as GPT-4 and Gemini design a custom study plan using the explanation based on the learning preferences and weaknesses of students. The learning preference includes simple explanations, practice problems, and quizzes that adjust based on the student’s progress to make learning easier and more personalized. Additionally, generating explanations using LLMs makes the learning process more understandable and interesting. Researchers \cite{abu2023building} explain that large language models provide graph-based explanations to organize the whole syllabus for students, and allow students to follow each step systematically for better understanding. With these organized step-by-step explanations, students will know why they study particular topics, thus making them interested.\\
\indent In \textbf{urban planning and smart city applications}, LLM explainability helps break down complex urban questions into smaller, manageable factors and provide detailed, clear answers. Take the case of traffic congestion arising in the city's downtown area. As described in \cite{jiang2024urbanllm} explanation, using LLMs breaks down traffic congestion issues into contributing factors by factors such as traffic flow patterns, road capacities, and peak travel times. Then, the language model provides detailed answers, such as optimizing traffic lights, creating bike lanes, or extending public transport routes, thus making the whole urban planning process more effective and efficient.
\section{Challenges and Limitations related to Explainable AI (XAI)}
\begin{figure*}[htbp]
    \centering
    \includegraphics[width=\linewidth]{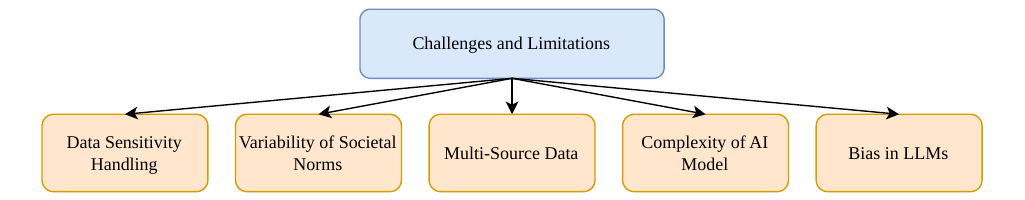} 
    \caption{Summary of the challenges of XAI}
    \label{fig:challenges}
\end{figure*}

Explainable AI (XAI) aims to improve AI-based decision-making processes by building user trust with fairness, accountability, and transparency. Significant challenges exist as shown in Figure \ref{fig:challenges}, especially within large language models (LLMs) for XAI, due to issues including 1) sensitive data handling, 2) accommodating societal diversity and norms, 3) the use of multisource data and algorithms, 4) complexity of AI model and 5) bias and fairness within LLMs
\subsection{Sensitive Data}
One of the key challenges in XAI is the limited access to sensitive data (e.g., personal or private information) required to explain certain decisions, especially in post-hoc approaches (Section \ref{sec:post_hoc}) and human-centered approaches (Section \ref{sec:human_centre}). Sensitive data include personal or private information, such as medical histories, financial transactions, or social media activity. For example, suppose that an AI model predicts that a patient in the ICU is at high risk of mortality due to sepsis (a severe and life-threatening condition caused by the body’s response to infection that damages its tissues and organs) \cite{Singer2016Sepsis3}. In the case of predicting diseases, such as sepsis, the decision may require disclosing specific details from the patient’s medical record, including the progression of vital signs, biomarkers (such as lactate levels), or infection history. While the disclosure of such records is necessary for clinical decision-making, it could risk the patient's privacy. \\
\indent Robust security systems are required to protect sensitive data from potential risks \cite{Farayola2024DataPrivacySecurity}, such as unauthorized access and data breaches. Failure to provide security systems would have serious repercussions \cite{Kuzniacki2022}, such as legal penalties, loss of user trust, financial losses, and reputational damage. XAI systems must establish clear guidelines on data handling, such as implementing data encryption, anonymizing sensitive information, and restricting access through authentication protocols \cite{jaff2024dataexposurellmapps, YAO2024100211}, to ensure reliability with privacy standards. XAI can help detect and mitigate security vulnerabilities, ensure that sensitive data is handled responsibly, and minimize risks such as unauthorized access to data by making operations of AI models transparent as discussed in section \ref{sec:intrinsic}.
\subsection{Societal Diversity and Norms}
Another significant challenge is accommodating societal diversity and varying norms in models for the XAI. Diversity varies across and within different groups \cite{anik2024diversity}, so it is important to design AI models that respect and reflect differences in diversity. For example, in medical science, an individual of African descent is more susceptible to sickle cell anemia due to genetic reasons \cite{williams2016sickle}, whereas an individual of European descent is more prone to cystic fibrosis \cite{cardoso2004low}. An AI system that predicts disease risk would need to take these demographic-dependent differences into account to ensure the results will be accurate and fair.\\
\indent Similarly, societal norms vary across cultures and communities \cite{gerston2010public}, so it is important to design AI models that carefully give attention to culturally specific norms and expectations to remove unintentional confusion or harm \cite{peters2024culturalbiasexplainableai}. For example, in the system of content moderation, what is considered appropriate or offensive can vary greatly between cultures. For example, the "thumbs up" gesture means "good job" or "I agree" in many Western countries. But in some Middle Eastern countries, it can be seen as rude or insulting. The above example highlights the need to understand and respect cultural differences in XAI systems.
\subsection{Multi-Source Data}
XAI also faces the challenge of combining multiple data sources. The final explanation may be skewed because AI systems use various types of data, such as social media activity, financial transactions, and online behavior, and each type of data provides a different perspective \cite{DEBRUIJN2022101666}. Using multi-source data and giving less importance than the required for optimal performance can distort the results \cite{masyutin2015credit}. For example, in a credit scoring system (a method used to evaluate someone's creditworthiness), if social media data, such as tweets, are given more importance than financial transaction history, the system might overestimate and underestimate the applicant's creditworthiness \cite{day2022relationship}. Misalignment occurs because social media data may lack the accuracy or direct relevance of financial transaction history in predicting credit behaviors. Consequently, the resulting credit score fails to predict and explain the financial stability of the applicant. In such cases, the real information needed to predict the score should come only from financial transaction data, making the use of social media data unnecessary or misleading. A similar issue occurs in stock prediction models, where predictions may be influenced by Twitter posts about rumors, even when the company's financial data suggest that the stock is stable. 
\subsection{Complexities of AI Model}
\indent Understanding the complexity of AI is very essential for the improvement in the explainability of the predictions \cite {complexity_ai} and fostering trust in the decision-making process of these AI models, both at the theoretical and application level. At a \textbf{theoretical level}, XAI emphasizes the importance of breaking down the decision-making process into simpler and more understandable parts. For example, in a neural network with five layers, the first layer might focus on recognizing basic features, such as edges and textures in an image. The second layer then combines simple features to detect more complex patterns, such as shapes and parts of objects. The difficulty is understanding how the decisions made in the first layer influence the decisions made in the second layer, and how each subsequent layer builds on the previous one. To fully understand the model, it’s essential to see how each layer works together to make the final decision or prediction \cite{rauker2023toward}.  The complexity of the AI model \cite{umer2023pyramidtabnet, mohsin2023zero} poses challenges not only at the theoretical level but also in real-world applications. Designing user interfaces that help people interact with and understand complex AI systems becomes an additional challenge \cite{nguyen2024human}. For example, interfaces can be implemented at different levels of the model. At the first layer, the interface could focus on explaining low-level features, such as edges or textures. As the model progresses through the layers, the interface would show how these low-level features are combined into more complex patterns, such as shapes and object parts, in intermediate layers. In the final layer, the interface would explain how these combined features lead to the model’s final decision or prediction. For the \textbf{application level}, it is important that interfaces cater to different users based on their level of expertise, autonomy, and trust in the AI system. For example, a beginner might need a simpler explanation of the basic features, while an expert might want more in-depth insights into how the model combines information across all layers. By addressing the complexity of AI model, we can provide clear, effective explanations tailored to each user’s needs \cite{Lim2009} \\
\indent Furthermore, after designing the interface, deploying complex AI models for predictions and utilizing LLMs for their explainability can be challenging. There are two main reasons for this: one is the limited power of the computers, such as GPUs needed to run complex AI models, and the other is the restrictions on smart devices, such as smartphones, which have less storage and processing ability. \cite{IBM2024}. For the first issue, computers may not have enough power or the right tools to run heavy AI models. The tools to run a heavy AI model can be expensive, and getting access to better computers might not always be possible. For the second issue, if edge devices, such as smartphones are used to reduce costs, the devices themselves often lack the storage space or processing power to handle large models locally. Techniques, such as Knowledge Distillation, have been developed to overcome the issue of limited processing and computation ability. Knowledge Distillation involves transferring knowledge from a large model  (e.g., BERT, which is a large and powerful language model used for understanding text) to a smaller model (e.g., DistilBERT, a smaller version of BERT that performs similarly but requires less power and resources). In Knowledge Distillation, the smaller model (DistilBERT) learns to mimic the performance of the larger model (BERT) while being more efficient. The same method, as highlighted by \cite{Cantini2024}, reduces the computational constraint of AI models, making them more suitable for resources-constrained environments, including XAI applications.
\subsection{Bias and Fairness within LLMs}
Another limitation is the bias and fairness issues presented within LLMs \cite{huang2023bias}. Addressing the biases and fairness in large language models (LLMs) is important to ensure that the outputs of AI model are equitable and accessible for diverse users with different contexts. Biases include 1) social, 2) language, and 3) representation biases in the responses. \\
\indent \textbf{Social biases} are often related to characteristics, such as age, gender, and lifestyle, impacting individuals across different social groups unequally, as noted in \cite{huang2023bias} and \cite{caliskan2023artificial}. For instance, it might recommend men for leadership positions, such as managers and executives more frequently than women, even when both have similar qualifications. Social bias stems from historical data that overrepresents men in such roles, perpetuating stereotypes and unfairly disadvantaging female candidates.\\ 
\indent \textbf{Language biases} emerge when LLMs treat different languages unequally, creating social disparities in areas, such as communication and education \cite{mohsin2025retrieval, recommend_biases}. For instance, the model might give better and more accurate responses in English than in less commonly spoken languages, which could make its outputs less useful or accessible to non-English speakers. Biases in LLMs-generated content, such as recommendation letters and news articles, occur due to language biases. Language biases can even affect translation tasks, potentially leading to misrepresentation or inaccessibility for different language groups. \\
\indent \textbf{Representation biases} occur when the data used to train the model lack balance, such as underrepresenting certain communities, cultures, and viewpoints. The skewed data can make the model generalize unfairly or completely overlook specific groups. For example, suppose an LLM is tasked with explaining a decision related to job recruitment but has been trained on data that overrepresent male candidates in leadership roles. In that case, its explanations might unintentionally favor male applicants or use language that perpetuates gender stereotypes. \\
\indent Biases in LLMs would increase the challenges in using LLMs for model explainability. For example, when explaining a recruitment decision, social biases might lead to stereotyped justifications favoring male candidates, language biases could make the explanation less comprehensible in non-English contexts, and representation biases might exclude perspectives of underrepresented groups, resulting in unfair and inaccessible explanations. Recent studies \cite{82ac6ae737e94dd6809b8d17b2621a68} reveal that even state-of-the-art LLMs, such as ChatGPT show unfairness in critical fields, such as education and healthcare. \\
\indent Researchers have proposed different methods, such as in-context learning \cite{fei2023mitigating} and refining prompt \cite{kamruzzaman2024prompting, memon2024llm}, for the problems of biases and fairness \cite{hua2024up5unbiasedfoundationmodel}. For example, if an AI is asked about a sensitive topic, providing some background about the topic can help the model respond more thoughtfully. Similarly, researchers might avoid asking a question about a person's age, gender, or ethnicity, or they might rephrase a question to make sure the answer doesn't favor one group over another.
\begin{figure*}[htbp]
    \centering
    \includegraphics[width=0.80\textwidth]{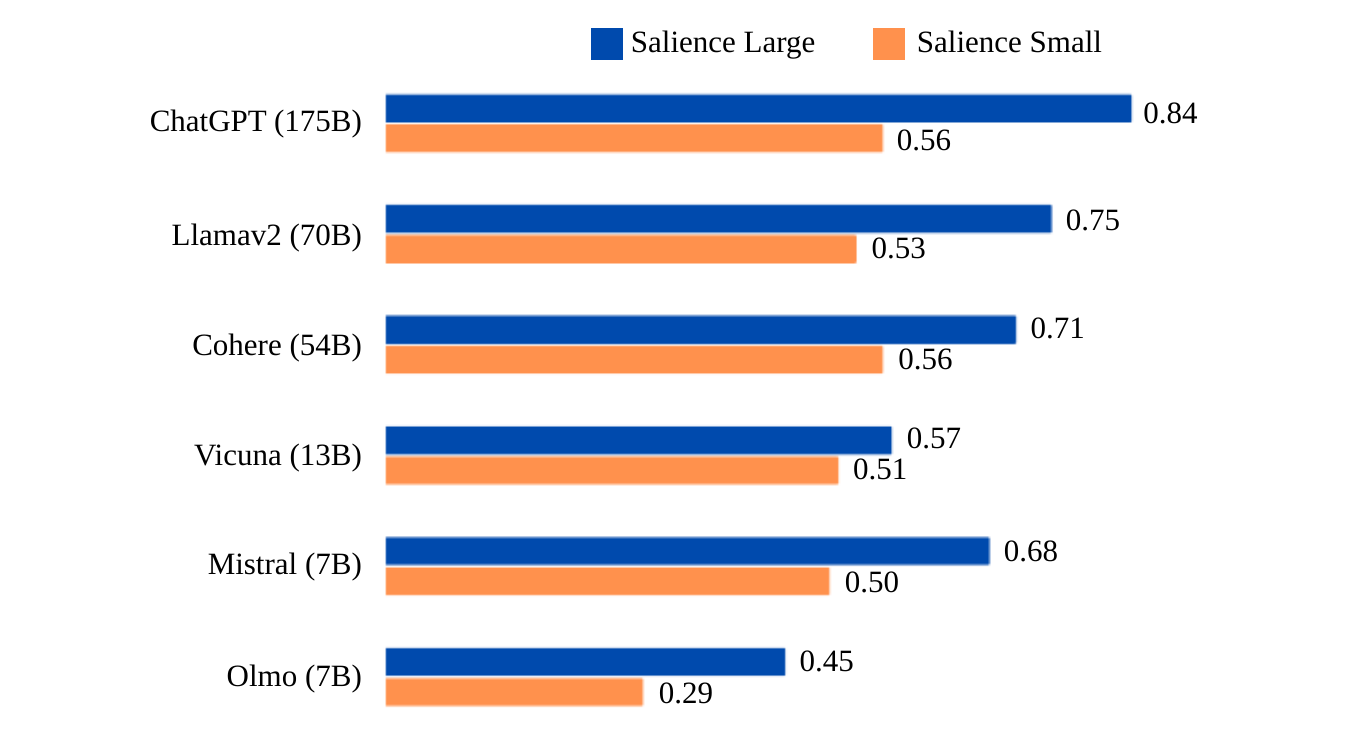} 
    \caption{Saliency Comparison of Models by \cite{koo2024benchmarking}}
    \label{fig:saliency}
\end{figure*}
\section{Feature Importance Across LLMs used for Explanation Using Saliency Maps}
The comparison of saliency values (see Figure \ref{fig:saliency} across various large language models (LLMs) highlights their emphasis on critical features, a key aspect of explainable AI. Saliency techniques help us identify the parts of a generated explanation, such as certain words and phrases that are most aligned with the ground truth \cite{CEREKCI2024111356}. The saliency techniques typically involve 1) gradient-based methods (which measure the change in the output by a small change in the input by identifying important features in the input), and 2) attention mechanisms (which allocate focus on the parts of the input that are most relevant to the model). The saliency techniques help to measure the importance of the features, such as specific work or phrases in the explanation of the decision or response generated by the large language models. \\
\indent Highlighting the important parts of the word or phrase offers a clearer view of the prioritization of features by different large language models during the explanation generation and the alignment of the generated response by the ground truth. For example, findings by \cite{koo2024benchmarking}, as shown in Figure \ref{fig:saliency}, provide saliency scores for each LLM under two conditions: one with a larger and one with a smaller input context. This is to ensure that LLMs maintain a consistent focus on important features regardless of the length of the provided input context. In this, ChatGPT (175 Billion parameters) \cite{chatgpt} has the highest saliency values, with 0.84 for large inputs and 0.56 for small inputs, indicating its strong ability to identify and prioritize important features. Similarly, LLAMAV2 (70B) \cite{touvron2023llama2} and COHERE (54B) \cite{almazrouei2023falcon40b} also show significant saliency values (0.75 and 0.71 for large inputs, respectively), highlighting their effectiveness in recognizing important elements. In contrast, models, such as VICUNA (13B) \cite{chiang2023vicuna} and MISTRAL (7B) \cite{jiang2023mistral} display lower saliency values, suggesting they place less emphasis on critical features, especially in smaller input contexts. OLMO (1B) \cite{groeneveld2024olmo} has the lowest saliency scores (0.45 for large inputs and 0.29 for small inputs). The above result is presented by \cite{koo2024benchmarking}, which indicates that the disparity among models emphasizes the importance of size, architecture, and training strategies in saliency performance.
\begin{figure*}[htbp]
    \centering
    \includegraphics[width=0.7\linewidth]{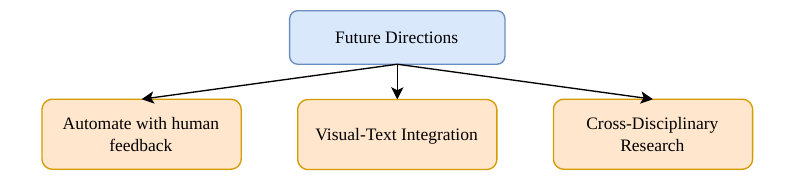} 
    \caption{Future directions for improving explainability in AI using LLMs}
    \label{fig:future_section}
\end{figure*}
\section{Future Directions}
In this survey paper, we discussed the importance of making AI more understandable, focusing on how large language models (LLMs) can help achieve this as shown in Figure \ref{fig:future_section}. We also explored the challenges and limitations of using LLMs for explainability, such as the complexity of their decision-making processes. This survey paper also discusses the importance of features, such as specific words or phrases, in the explanations generated by large language models using saliency maps. \\
\indent Future directions will focus on 1) automating feedback, 2) linking text with visuals, and 3) fostering collaboration across different fields to make AI explanations clearer and more helpful. \textbf{Automating human feedback} would enable AI to adapt its explanations based on user interactions, thus saving time and effort. For example, a medical AI would fine-tuning the explanations by learning from feedback received from doctors \cite{aftabxai}. Similarly, including \textbf{text-based explanations combined with visuals}, such as heatmaps \cite{Selvaraju_2019,kim2018interpretabilityfeatureattributionquantitative} would explain AI decisions, such as showing highlighted areas in a medical image alongside a clear written explanation. \textbf{Cross-disciplinary collaboration} allows experts in AI, cognitive science, and domain-specific regions, such as medicine and law to come together with the goal of instinctive, relevant, and actionable explanations. For example, such collaboration can  allow medicine to create AI capable of explaining its diagnostic decisions in a understandable way by doctors and patients, so they could act upon those findings. The above-discussed efforts thus make AI even more practical and trustworthy for a variety of real-world applications. 
\subsection{Automation with Human Feedback for XAI Applications}
Explainable AI (XAI) involves methods that help users understand and trust the decisions made by AI models, as discussed earlier in section \ref{sec:intro}. Previous section \ref{sec:human_centre} introduces human feedback through surveys and interviews to refine AI-generated explanations, making them easier to understand \cite{do2024facilitatinghumanllmcollaborationfactuality}. However, relying solely on manual feedback collection can be time-consuming and impractical, especially for large-scale AI applications. For example, in a customer support chatbot used by an e-commerce site with millions of users daily. Manually collecting feedback and analyzing all the responses would take too much time and effort. The future direction is to automate human feedback with AI processes to make the system faster and more efficient. For example, AI could track patterns in user interactions with explanations, such as noticing which explanations are useful and where they often ask for clarification. The automated system can use the patterns to quickly adjust the explanations and enable the explanations to match users' needs without requiring continuous manual assistance from the people. One practical example of an automatic response system is a movie recommendation system \cite{reddy2019content}. Suppose a platform that explains the suggestions of the recommendation system by saying: 'You liked action movies starring Actor X' If users find this explanation unclear or unhelpful, the system could adapt by providing simpler language or additional details, such as mentioning similar movies. Automation would save time and make AI interactions smoother and more practical. \\
\indent The approach of automating human feedback to XAI systems can be applied across various sectors, such as healthcare and finance, to enhance the practicality and efficiency of AI systems. In healthcare, suppose that an AI tool assists doctors in disease diagnosis \cite{aftabxai}. The tool explains its decision, such as "This diagnosis is based on symptoms X, Y, and Z observed in the patient’s history." If doctors find the explanation lacking information or unclear, they might manually flag it. Over time, an automated feedback system could analyze common feedback patterns and adjust its explanations, such as including references to recent medical research or visualizing patient data trends for clarity. For the finance sector, in a financial advising app, the system might explain investment recommendations, such as "This stock is suggested due to its high growth rate and low market volatility (market uncertainty)"\cite{bianchi2021robo}.  If users frequently request clarification on terms, such as "market volatility," the AI could adapt by adding definitions or visual aids, such as charts showing the stock's historical performance. In the education sector \cite{manna2024need}, an AI tutor helping students with math problems could explain solutions in steps, such as "Step 1: Combine alike terms. Step 2: Simplify the equation by performing basic operations to make it easier to solve. Step 3: Solve for the unknown by separating the variable." If many students indicate confusion, the AI could refine its explanation by adding examples, such as "For instance, 3x + 2x becomes 5x because you add the coefficients." In this way, AI systems can meet users' needs in each field.
\subsection{Integrating Visual and Text Explanations} \label{sec:fut_visualize}
To enhance the explainability of AI models, future efforts could focus on combining textual and visual explanations. For example, when a model predicts a specific outcome, a corresponding text-based explanation can be presented alongside visual aids, such as heatmaps \cite{lee2022heatmap}, to show the parts of the input that influence the decision. This dual modality (textual response with visual explanation) can make it easier for users to understand the model's reasoning by combining textual explanations with visual aids. The text provides clarity, while visual aids complement it by highlighting key input features that influence the output, making the connection between inputs and outputs more accessible. \\
\indent One approach involves integrating text explanations from LLMs with visualization techniques, such as Grad-CAM (Gradient-weighted Class Activation Mapping) \cite{Selvaraju_2019}. For instance, suppose that an AI model diagnoses medical images: Grad-CAM can highlight the regions on an X-ray that contributed to the diagnosis \cite{LAMBA2024110159}, while an LLM provides a text explanation, such as, "The highlighted region shows abnormalities consistent with pneumonia." This combination helps users understand the diagnosis both visually and contextually. \\
\indent A promising future direction is using knowledge graphs to connect text-based explanations from LLMs with visual representations. Knowledge graphs help organize and display relationships between important pieces of information to make complex AI outputs easier. For example, in an LLM chatbot, a knowledge graph could show how a user's question is linked to the information retrieval system and the response it generates.
\subsection{Cross-Disciplinary Research}
Explainable AI (XAI) is inherently a multidisciplinary effort \cite{LONGO2024102301}, for example collaboration between researchers from fields, such as computer science, cognitive science (which focuses on understanding how humans perceive and interpret information), human-computer interaction (HCI) (which studies the design and use of computer technologies about user experience), and domain-specific areas (e.g., healthcare or finance) can be envisioned for the effective use of AI \cite{Langer2021}. Each field offers unique perspectives, making it difficult to agree on shared design principles or explainability evaluation standards \cite{Speith2022}. The main reason for this is that the number of publications in XAI is growing so quickly that researchers struggle to keep up with studies in their field. This leaves them with little time to explore work from other fields, which also has a large and overwhelming amount of research. \\
\indent To address this, collaboration between AI researchers, cognitive scientists, and domain experts is essential to advance the use of large language models (LLMs) in explainable AI (XAI) \cite{Cabour2023, LONGO2024102301}. Cognitive scientists, who study how the human brain processes information, can help design LLM-generated explanations that align with how people process information, ensuring they are clear and intuitive for users. Domain experts, such as doctors or lawyers, can guide the development of explanations that are tailored to the professional requirement, making them both relevant and actionable. \\
\indent For example, collaboration between AI researchers and healthcare professionals could lead to LLM-based explainability techniques designed for medical decision-making. LLMs could generate textual explanations that summarize patient data or medical imaging results in ways that are understandable to clinicians \cite{leeftink2020multi}. Cognitive scientists could further enhance explanations by studying how doctors interpret visual aids, such as Grad-CAM heatmaps as discussed in section \ref{sec:fut_visualize}, helping to integrate textual and visual explanations into a seamless and user-friendly format \cite{LAMBA2024110159}. This approach could pave the way for more effective and interpretable AI systems in specialized fields.

\bibliographystyle{ACM-Reference-Format}
\bibliography{main}         
\end{document}